\documentclass[a4paper,fleqn]{cas-dc}

\usepackage[numbers]{natbib}

\usepackage{multirow}
\usepackage{array}
\usepackage{amsmath}
\usepackage{amssymb}
\usepackage{booktabs}
\usepackage{color}

\usepackage{lineno,hyperref}

\usepackage[colorlinks]{hyperref}

\def\tsc#1{\csdef{#1}{\textsc{\lowercase{#1}}\xspace}}
\tsc{WGM}
\tsc{QE}

\begin{document}
\let\printorcid\relax
\let\WriteBookmarks\relax
\def\floatpagepagefraction{1}
\def\textpagefraction{.001}
\title {Multi-modal and Multi-view Fundus Image Fusion for Retinopathy Diagnosis via Multi-scale Cross-attention and Shifted Window Self-attention} 
\author[a,b]{Yonghao Huang}
\author[a,c]{Leiting Chen}
\author[a,c]{Chuan Zhou}
\corref{mycorrespondingauthor}
\ead{zhouchuan@uestc.edu.cn}
\cortext[mycorrespondingauthor]{Corresponding author}
\address[a]{Key Laboratory of Intelligent Digital Media Technology of Sichuan Province, University of Electronic Science and Technology of China, Chengdu, Sichuan Province, China}
\address[b]{School of Computer Science and Engineering, University of Electronic Science and Technology of China, Chengdu, Sichuan Province, China}
\address[c]{School of Information and Software Engineering, University of Electronic Science and Technology of China, Chengdu, Sichuan Province, China}
\date{}

\begin{abstract}
    The joint interpretation of multi-modal and multi-view fundus images is critical for retinopathy prevention, as different views can show the complete 3D eyeball field and different modalities can provide complementary lesion areas. 
    Compared with single images, the sequence relationships in multi-modal and multi-view fundus images contain long-range dependencies in lesion features. 
    By modeling the long-range dependencies in these sequences, lesion areas can be more comprehensively mined, and modality-specific lesions can be detected. 
    To learn the long-range dependency relationship and fuse complementary multi-scale lesion features between different fundus modalities, 
    we design a multi-modal fundus image fusion method based on multi-scale cross-attention, which solves the static receptive field problem in previous multi-modal medical fusion methods based on attention. 
    To capture multi-view relative positional relationships between different views and fuse comprehensive lesion features between different views, we design a multi-view fundus image fusion method based on shifted window self-attention, which also solves the computational complexity of the multi-view fundus fusion method based on self-attention is quadratic to the size and number of multi-view fundus images. 
    Finally, we design a multi-task retinopathy diagnosis framework to help ophthalmologists reduce workload and improve diagnostic accuracy by combining the proposed two fusion methods. 
    The experimental results of retinopathy classification and report generation tasks indicate our method's potential to improve the efficiency and reliability of retinopathy diagnosis in clinical practice, achieving a classification accuracy of 82.53\% and a report generation BlEU-1 of 0.543. 
    
\end{abstract}


\maketitle

\section{Introduction}
\label{sec:introduction}

\begin{figure}[htb]
    \begin{center}
        \includegraphics[width=0.49\textwidth]{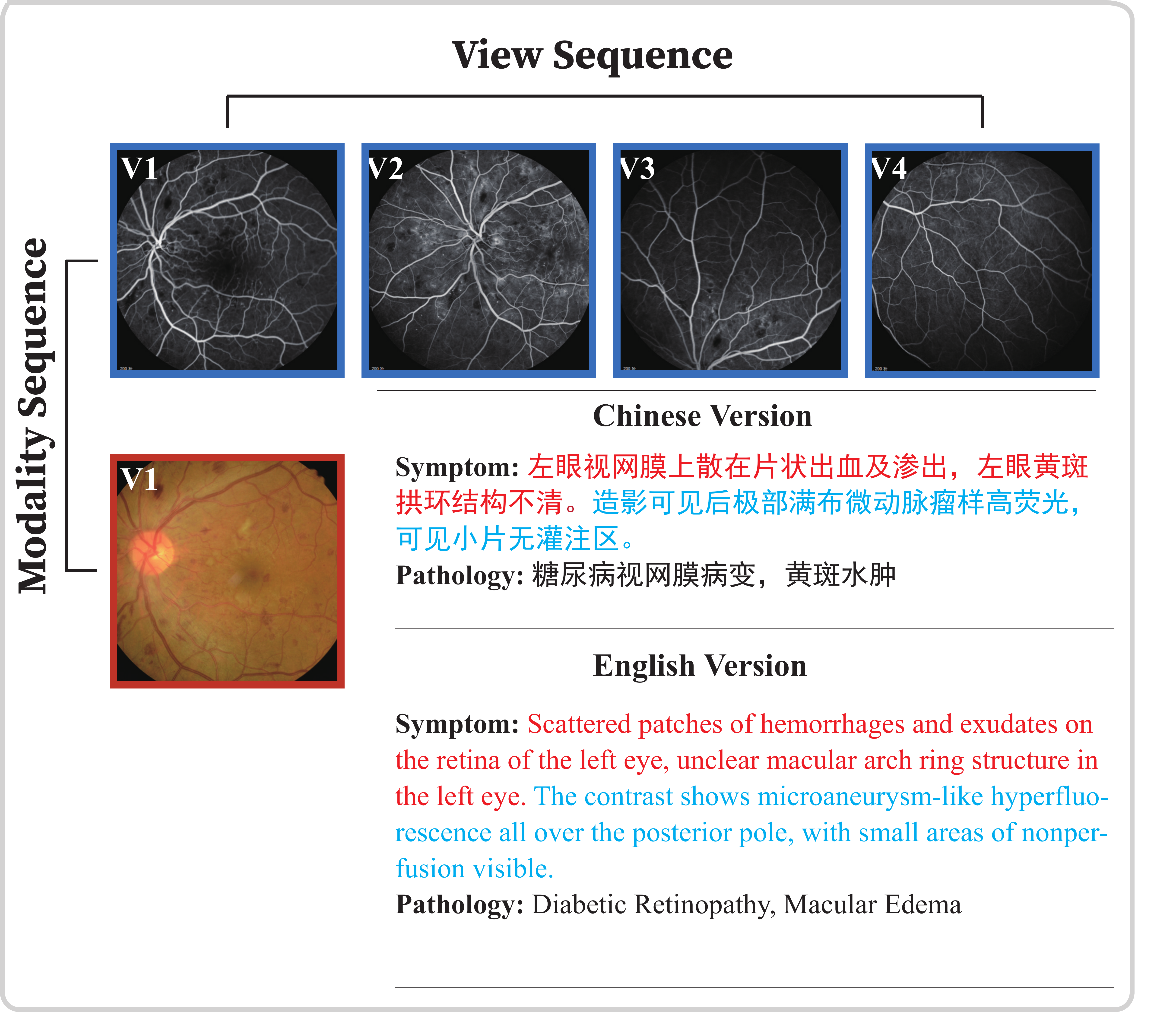}
        \caption{An example of multi-modal and multi-view fundus images with symptom reports. The diagnostic report is based on five fundus images with two modalities, where the top four images (Vl, V2, V3, V4) are FFA images with different views, and the image in the bottom left corner is a CFP image. Each report contains symptom and pathology sections, where the phrases marked in red are observations from the CFP image, and the phrases marked in blue are observations from these FFA images. Multi-modal and multi-view fundus images contain modality sequence relationships and view sequence relationships, compared with single-modal images. }
        \label{figure_introduction1}
    \end{center}
\end{figure}

\begin{figure}[htb]
    \begin{center}
        \includegraphics[width=0.49\textwidth]{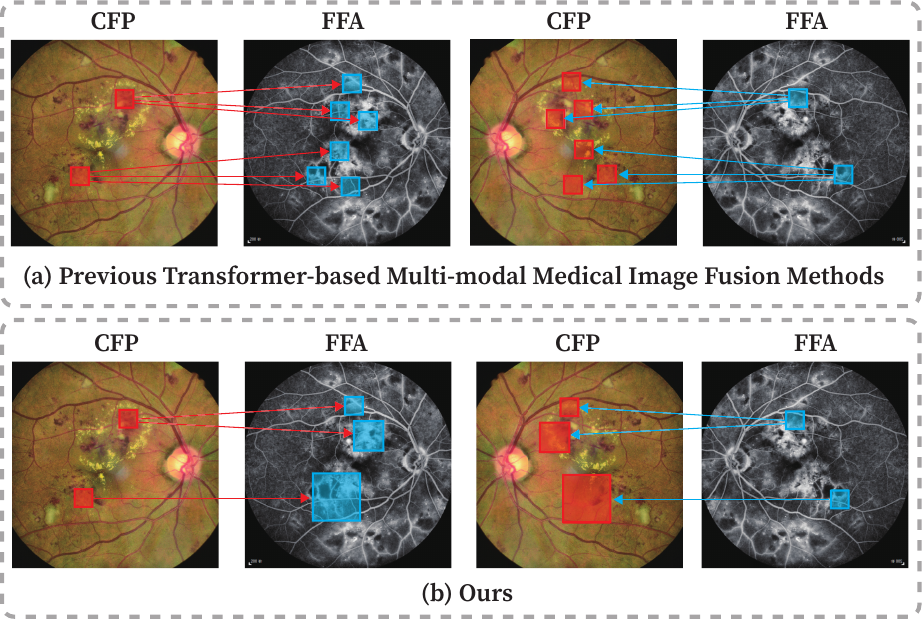}
        \caption{Comparison of different attention mechanisms in previous multi-modal medical image fusion methods and our CFFT. The size of the box indicates the receptive field size of the corresponding token embedding. These arrows indicate the correspondence relationships between inter-modal token embeddings. }
        \label{figure_introduction3}
    \end{center}
\end{figure}

Retinopathy (e.g., cataract, glaucoma, and diabetic retinopathy), responsible for most worldwide vision impairment, will cause more than 61 million people to become blind by 2025 \cite{burton2021lancet}. 
Fundus images, such as color fundus photography (CFP) images and fundus fluorescein angiography (FFA) images \cite{lee2019primary}, are widely used in clinical settings for the preventive, diagnostic, treatment, and rehabilitative interventions of retinopathy. 
Thus, plenty of automated methods based on fundus images \cite{kallel2024retinal,sivapriya2024automated,yu2021mil,zhou2021automatic,qureshi2021diabetic} have been proposed to assist retinopathy diagnosis and treatment. 
However, these methods usually rely on only a single fundus image as the input to train models for retinopathy automated diagnosis tasks. 
In clinical practice, ophthalmologists usually write the symptom sections and pathology results in fundus symptom paragraphs based on the multi-modal and multi-view fundus images since different views can show the complete eyeball field and different modalities can provide complementary lesion information, as shown in \ref{figure_introduction1}. 
A single fundus image can hardly record complete fundus information, because it can only provide a fundus field of view of 45° \cite{takahashi2017applying}. 
Additionally, it is difficult for a single-modal fundus image to provide comprehensive information on the lesion's appearance for ophthalmologists since different modal fundus images show distinct symptoms of specific retinopathies. 
For example, the CFP image is a non-invasive fundus image \cite{li2021applications} taken directly with a monocular camera that provides a clear picture of some of the important biomarkers of the fundus, such as the optic cup, macula, and optic disc. 
FFA is a fundus screening modality that records information about the fundus retina by performing luminescence with fluorescein \cite{lee2019primary}, which not only highlights the filling defect of the fundus blood vessels but also allows abnormalities in the retinal pigment epithelium and choroidal vessels to be observed through fluorescent performance. 
These CFP and FFA findings can help ophthalmologists to diagnose and treat retinopathy. 

The excellent performance of convolutional neural networks (CNNs) in computer vision shows that they can facilitate the diagnosis of retinopathy based on fundus images. 
Recently, some studies \cite{quan2025multi,krishna2024optimizing,el2024multimodality, MVDRNet,lan2021automatic,wang2019two} proposed multi-modal and multi-view fusion methods based on CNNs to properly combine features of different views and different modalities for improving the accuracy of automated methods. 
However, the constrained local receptive fields of convolution make it difficult for these methods to learn long-range dependencies in multi-modal and multi-view fundus images. 
Compared to the single fundus image, there are many sequence relationships in multi-modal and multi-view fundus images, as shown in \ref{figure_introduction1}. 
By modeling the long-range dependencies in these sequences, lesion areas can be more comprehensively mined, and modality-specific lesions can be detected. 
For example, microaneurysm \cite{li2021applications} in CFP images and neovascularization \cite{lee2019primary} in FFA can be detected by modeling the long-range dependencies in the modality sequences. 

On the one hand, some methods \cite{Transmed,huang2023model,xiao2024cross,li2024transiam, bi2024multimodal} proposed various multi-modal image fusion strategies based on self-attention and cross-attention to model long-range dependency among multi-modal medical images and fusion multi-modal information for ancillary diagnostic tasks. 
For instance, some multi-modal fusion strategies based on self-attention \cite{Transmed,huang2023model} for mining long-range dependent information in multi-modal magnetic resonance imaging images and multi-modal fundus images, respectively. Xiao et al. \cite{xiao2024cross} proposed a multi-modal fundus image fusion method based on cross-attention to capture the correspondence between CFP images and infrared fundus photography images for retinopathy grading. 
Nevertheless, these methods have largely ignored the multi-scale lesion objects because of the static receptive fields of token embeddings in the regular self-attention and cross-attention, making them brittle to clinical practices that involve lesion objects of distinct sizes in different modalities. 
Inspired by the shunted self-attention \cite{Shunted} with multi-scale receptive, 
we extend it and design a multi-scale cross-attention, which denotes multi-scale cross-fundus attention (MCA), to learn the correspondence between different fundus modalities through the mutual attention mechanism of cross-attention. 
To capture and fuse more multi-scale fundus information from different modal fundus images, we design a multi-modal fundus fusion method, which denotes a cross-modal fundus fusion transformer (CFFT), by stacking multiple MCA blocks with different receptive fields. 
In Figure \ref{figure_introduction3}, we show a qualitative comparison between previous transformer-based multi-modal medical image fusion methods and CFFT. 
Previous methods conduct the long-range dependency relationship among multi-modal fundus images, yet in a static scale receptive field (Figure \ref{figure_introduction3} (a)). 
By contrast, the proposed CFFT captures multi-scale fundus lesion features yet simultaneously models the long-range dependency relationships among multi-modal fundus images (Figure \ref{figure_introduction3} (b)). 

On the other hand, the approach from \cite{huang2023model} models long-range dependencies among multi-view fundus images by concatenating token embeddings of different fundus views and computing the self-attention weights at the concatenated token embeddings. 
Nevertheless, the approach does not take into account the complementarity of position information between different fundus views. 
Therefore, we introduce a multi-view concatenate fusion strategy with multi-view fundus position embeddings to provide absolute position awareness and capture relative positional relationships between different views. 
To mine the long-range dependent information in a wide multi-view fundus field and fuse comprehensive lesion features between different views, we furthermore design a multi-view fundus shifted window fusion module (MFSWFM), which also solves the computational complexity of the multi-view fundus fusion method based on self-attention is quadratic to the size and number of multi-view fundus image. 

Finally, we design a multi-task retinopathy diagnosis framework (MRDF) to help ophthalmologists reduce workload and improve diagnostic accuracy by simultaneously processing multiple views and modalities of fundus images for various retinopathy diagnosis tasks. 
The proposed multi-task framework combines the MFSWFM and the CFFT as the encoder to fuse information of different modalities and views fundus images, and we employ a classifier and a recurrent long short-term memory (Re-LSTM) as the multi-task decoder to decoding the fusion multi-modal and multi-view fundus visual represents from the encoder for fundus retinopathy classification and fundus symptom report generation tasks. 
To verify the effectiveness and generalization of the proposed framework, we conduct experiments on two multi-modal and multi-view fundus image datasets, and three retinopathy diagnosis tasks. 
Experimental results indicate that the proposed framework demonstrates superior performance in various retinopathy diagnosis tasks by fully using fundus information from different modalities and views. 
In summary, our main contributions are presented as follows: 
\begin{itemize} 
\item To learn the long-range dependency relationship and fuse complementary multi-scale lesion features between different fundus modalities, we design a multi-modal fundus image fusion method based on multi-scale cross-attention, which solves the static receptive field problem in previous multi-modal medical fusion methods based on attention. 
MCA unifies multi-modal multi-scale fundus features interaction within one cross-attention layer via mutual attention and token aggregation mechanisms. 
\item To mine the long-range dependent information in a wide multi-view fundus field and fuse comprehensive lesion features between different views, we design a multi-view fundus image fusion method based on shifted window self-attention, which solves the computational complexity of the multi-view fundus fusion method based on self-attention is quadratic to the image size and number of multi-view fundus. 
We also introduce a multi-view concatenate fusion strategy and multi-view fundus position embeddings to provide absolute position awareness and capture relative positional relationships between different views. 
\item We demonstrate the superiority of our proposed CFFT and MFSWFM over previous multi-modal and multi-view medical image fusion strategies on three retinopathy diagnosis tasks. 
The results show the potential of multi-modal and multi-view fusion methods based on shifted window self-attention and multi-scale cross-attention for retinopathy diagnosis. 
\end{itemize}

\section{Related work}
\label{related work}
\subsection{Multi-modal medical image fusion}
Since medical images in different modalities often differ, effectively fusing information from different modalities to improve the performance of the model has become a problem of great concern in the field of multi-modal medical image processing. 
According to the fusion mode, multi-modal fusion methods based on deep learning can be categorized into three strategies: input-level fusion, layer-level fusion, and decision-level \cite{li2024review,zhou2019review}. 
These fusion methods were widely employed in various multi-modal medical image analysis tasks. 

In the input-level fusion, the different modal feature information fusion phase precedes the deep learning models by pixel merging and channel concatenation. 
The input level fusion is the most common in the field of multi-modal medical image analysis, widely used in various multi-modal medical image-assisted diagnostic tasks \cite{shaker2024unetr++,rallabandi2023deep,zhou2023nnformer}. The advantage of the input-level fusion is that the algorithm has a simple idea and can effectively preserve the spatial position information in the original image. The disadvantage is that it is difficult to establish the intrinsic connection between different modalities of medical images, resulting in a decrease in the performance of the algorithm. 
According to the related works \cite{li2024review}, layer-level fusion fusion multi-modal medical image features in intermediate layers of networks and can be divided into single-level fusion, hierarchical fusion, and attention-based fusion. 
The single-level fusion employs different backbones to extract features from different modalities separately, and then design fusion methods to fuse features and feed them to the final decision layer for various multi-modal medical image analysis tasks, such as autism spectrum disorders diagnosis based on multi-modal magnetic resonance imaging (MRI) images \cite{saponaro2024deep}, and Alzheimer's diagnosis based on multi-modal MRI and positron emission tomography images \cite{kadri2023efficient}, the detection of glaucomatous optic neuropathy based on visual Fields and peripapillary circular optical coherence tomography scans \cite{xiong2022multimodal}. 
The hierarchical fusion further extends single-level fusion by adding multiple parallel fusion modules in the middle layer of the network \cite{miao2024mmtfn, xu2023multi, tu2024multimodal}. 
As self-attention mechanisms \cite{li2024transiam, bi2024multimodal, chen2023multimodal} have been proposed and developed, more and more attention-based fusion methods are beginning to be proposed to utilize the attention mechanism for multi-modal information interaction and fusion. 
In the decision-level fusion strategy, the features of different modalities are extracted by different deep learning backbones and used to generate separate decision results for each modal. 
The final diagnosis results are produced by employing a fusion method, such as averaging or majority voting \cite{li2024review}, in these decision results. 
Nevertheless, these methods have largely ignored the multi-scale lesion objects as the static receptive fields of token embeddings, making them brittle to clinical practices that involve lesion objects of distinct sizes in different modalities. 
In this paper, we propose an attention-based multi-modal fusion method, which utilizes a multi-scale cross-attention to fuse coarse-grained and fine-grained lesion information and learn the correspondence long-range dependency relationship among different modalities. 

\subsection{Multi-view medical image analysis}
Combining medical images of different views can provide a wider field of vision and more feature information than single-view images. 
Therefore, research has been actively proposing various multi-view medical image analysis and fusion methods \cite{ming2024multi,luo2024lesion,huang2023model, MVDRNet,lan2021automatic,liu2021act,sun2019multi} to increase the performance of computer-aided diagnosis systems. 

To take advantage of the relationships of multi-view fundus images, Luo et al. \cite{MVDRNet} introduced the channel mechanisms to model the interdependencies between different views for improving the performance of diabetic retinopathy classification. 
Inspired by the sphere structure of the eyeball, Lan et al. \cite{lan2021automatic} employed a 3D convolution layer to fuse fundus features of different views for retinopathy symptom report generation tasks. 
However, these CNN-based multi-view fusion methods cannot model long-range dependency between different views. 
Ming et al. \cite{ming2024multi} employed a transformer as the backbone to extract long-range information and proposed a cross-view attention block for fusing information of craniocaudal and mediolateral oblique views. 
Recently, a multi-view fusion method based on self-attention \cite{huang2023model} was proposed to model the intra-view and inter-view long-range dependency, and the performance of retinopathy diagnosis systems achieves large improvement. 
However, the computational complexity of self-attention mechanisms is quadratic to multi-view fundus image size. 
Therefore, we propose a multi-view fusion method based on the shifted window self-attention, which achieves the linear computational complexity, to compute self-attention locally within multi-view shifted windows. 

\subsection{Vision transformers}
CNNs generally exhibit limitations for modeling global contexts, due to the convolution operations being fixed and attending to only a local subset of pixels in images. 
Inspired by the powerful global contextual modeling ability of the self-attention mechanism in transformers \cite{vaswani2017attention}, several researchers \cite{wu2021cvt, zheng2021rethinking} proposed hybrid CNN-Transformer models by introducing self-attention mechanisms into deep convolutional layers for various computer vision tasks. 
Dosovitskiy et al. \cite{Vit} applied self-attention mechanisms in shadow layers to construct a pure transformer-based network called Vision Transformer(ViT). 
Subsequently, many efficient ViT variants \cite{Shunted,kim2024learning,su2024can,Nat,Vig}, were proposed to improve the feature extracting ability of models via revising the tokens calculation operations in self-attention mechanisms. 
Since the above works demonstrated the ability of transformers to solve computer vision tasks, transformers have begun to be employed to explore the potential in various medical image processing tasks \cite{xiao2024cross,chen2021vit,zhang2022mmformer,huang2023model}. 
Chen et al. \cite{chen2021vit} combined ViT and CNNs to learn the long-range relationships between points in images for self-supervised volumetric image registration. 
Zhang et al. \cite{zhang2022mmformer} proposed multiple hybrid CNN-Transformer modal-specific encoders and a transformer-based modal-correlated encoder for multi-modal brain tumor segmentation. 
Huang et al. \cite{huang2023model} and Xiao et al. \cite{xiao2024cross} employed transformers to model intra-modal long-range dependencies and aligned inter-modal long-range correlations for retinal classification tasks. 
Li et al. \cite{li2024transiam} proposed a dual path branch segmentation model called TranSiam, which uses a position-aware aggregation module based on self-attention mechanisms in the middle layer of the network to fuse information between different modal features for multi-modal medical image segmentation. 
Our work explores the adaptation of a Transformer for multi-modal and multi-view fundus image fusion and is the derivation of the above works in multi-modal and multi-view medical image processing fields. 

\section{Methods}\label{methods}

\begin{figure*}[htb]
    \begin{center}
        \includegraphics[width=0.85\textwidth]{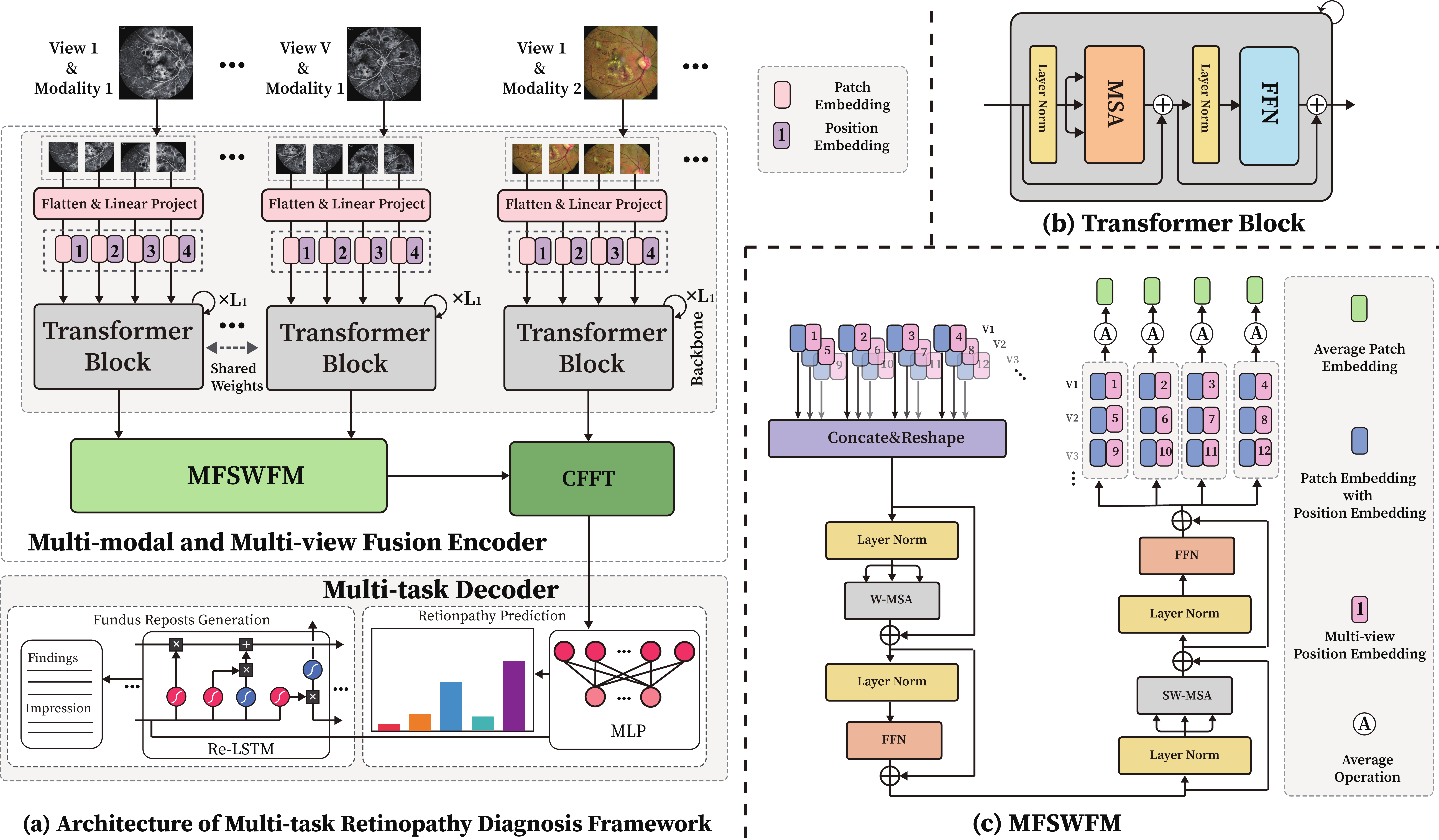}
        \caption{(a) The architecture of the proposed MRDF. The proposed framework is a network structure that can handle any number of modalities and views by expanding the backbone. For simplicity, only two modalities and two perspectives are shown; (b) The architecture of Transformer; (c) The architecture of MFSWFM. }
        \label{Methods}
    \end{center}
\end{figure*}

\begin{figure}[htb]
    \begin{center}
        \includegraphics[width=0.35\textwidth]{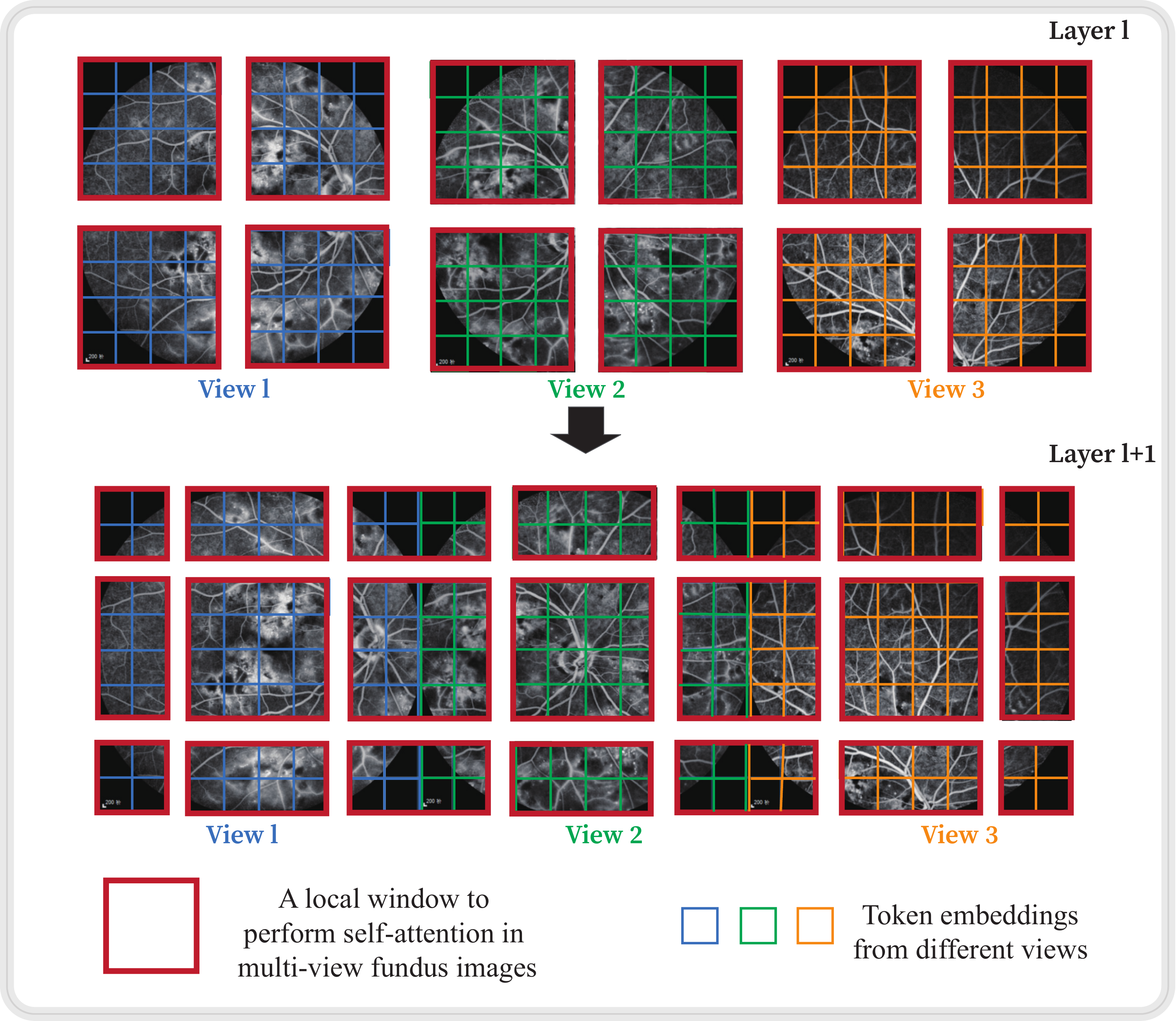}
        \caption{An illustration of the multi-view shifted operation for computing self-attention in the proposed MFSWFM architecture. In layer $l$ (up), these local windows with self-attention are applied in each view. In the next $l + 1$ (down), the window partitioning is shifted and bridges the token embedding of different views, providing connections among different views. }
        \label{figure_introduction2}
    \end{center}
\end{figure}

\begin{figure*}[htb]
    \begin{center}
        \includegraphics[width=0.80\textwidth]{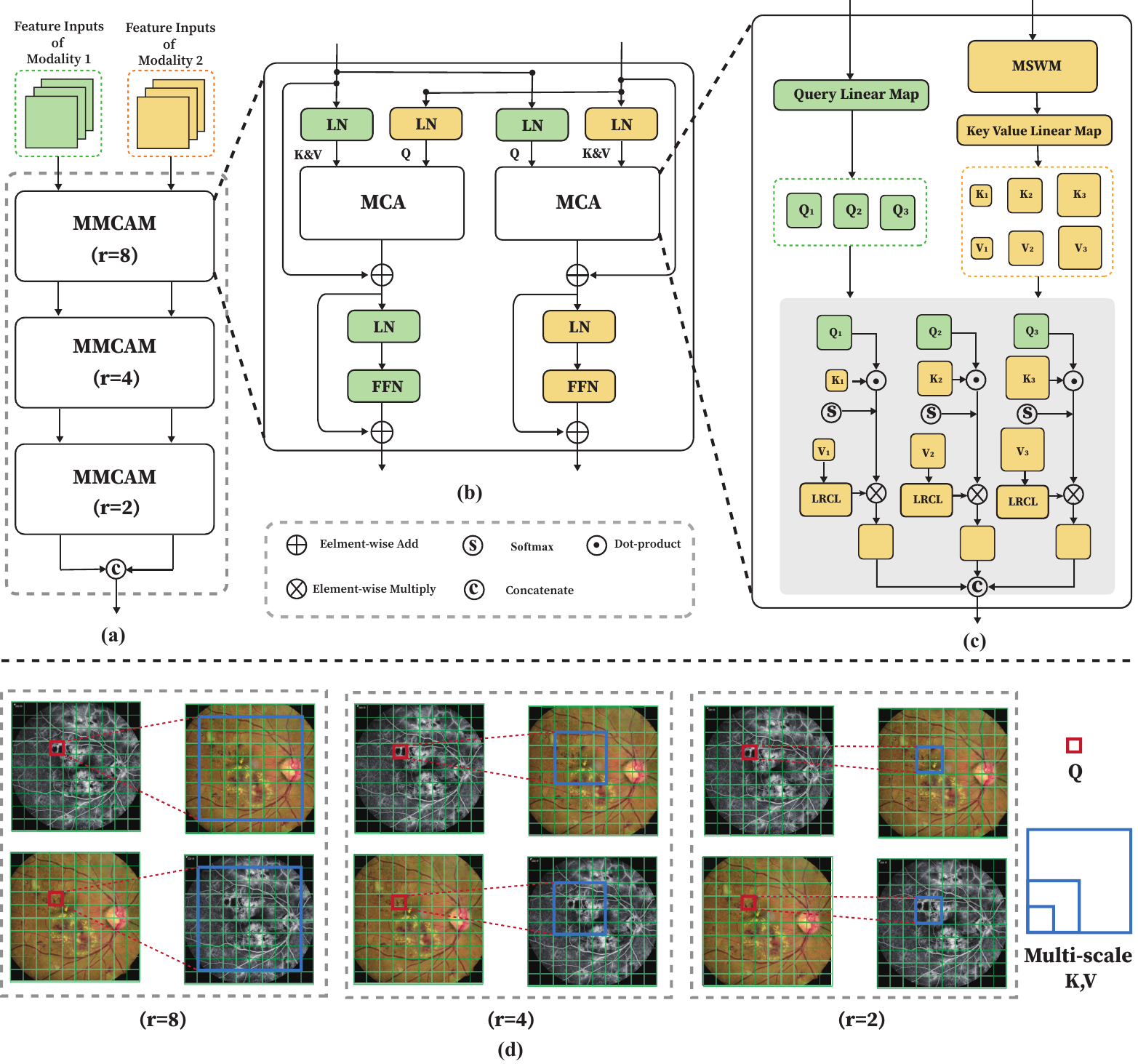}
        \caption{(a) The architecture of our proposed CFFT is presented, which contains three MMCAM parts with different parameters $r$; (b) The architecture of MMCAM; (c) The architecture of MCA; (d) A schematic diagram of receptive fields with different parameters $r$ in the proposed MMCAM. }
        \label{Methods2}
    \end{center}
\end{figure*}


\subsection{Framework and backbone}
The overall architecture of MRDF is presented in \ref{Methods}(a), which maintains a multi-task encoder-decoder architecture. 
Specifically, the encoder consists of three main components: (1) the backbone that contains multiple flatten and linear project operations, and multiple transformer blocks for extracting long-range dependency information in intra-modal and intra-view; (2) the proposed MFSWFM for extracting long-range dependency information in inter-view and fusing multi-view fundus features; (3) the proposed CFFT for extracting long-rang dependency information in inter-modal and fusing multi-scale features in multi-modal fundus images. 
Subsequently, the encoded feature vectors from the encoder are decoded by a multi-task decoder consisting of a classifier and a Re-LSTM for downstream retinopathy diagnosis tasks. 
The proposed framework can handle any number of modalities and views by expanding the transformer block in the backbone. 
For simplicity, we introduce the proposed framework by using two modal fundus images (CFP and FFA images) and FFA images of multiple views. 

Given a group of FFA images with different views $X_{FFA} =\left\{\mathbf{X}^{v} \in \mathbb{R}^{H \times W \times C} \mid v=1,2, \cdots, V\right\}$, where $(H, W)$ is the resolution of the image, $C$ is the number of channels, $V$ is the number of views in a sample, and $v$ is the index of the $v$-th view. 
Since the transformer-based architecture processes the embeddings in a sequence-to-sequence manner, we uniformly split the fundus image of each view into patch sequences. 
Then, we employ a flattening operation and a linear project to produce the local feature embeddings $\mathbf{X}_{emb}^{v} \in \mathbb{R}^{N \times D} $, where $D=(P^2 \cdot C)$ is the dimensions of embeddings, $P \times P$ is the resolution of each patch, and $N = \frac{HW}{P^2}$ is the number of embeddings. 
However, the flattening operation inevitably collapses the intra-view spatial information, which is critical to retinopathy diagnosis, in the fundus. 
To address this issue, we add a learnable position embedding $\mathbf{X}_{pos}^{v} \in \mathbb{R}^{N \times D}$ into $\mathbf{X}_{emb}^{v}$ to retain the spatial information of fundus images. We obtain the token embedding of the $v$-th view $\mathbf{X}_{0}^{v} \in \mathbb{R}^{N \times D}$: 
\begin{flalign}
\begin{aligned}\label{eq1}
    \mathbf{X}_{0}^{v} = \mathbf{X}_{emb}^{v} + \mathbf{X}_{pos}^{v} 
\end{aligned} 
\end{flalign}

Similarly, we can obtain the token embedding $\mathbf{Y}_{0}^{v} \in \mathbb{R}^{N \times D}$ of a CFP image following the above operations. 
We separately employ multiple transformer blocks based on ViT \cite{Vit} as backbone networks to extract intra-view and intra-modal features and model intra-view and intra-modal long-range dependencies. 
A transformer block consists of three main modules: Layer Normalization (LN), Multi-head Self-attention (MSA) and Feed Forward Network (FFN), as shown in \ref{Methods}(b). 
Specifically, we construct a transformer block with $L_{1}$ layers for each token embedding. 
To balance the performance and complexity of the model, we shared the parameters of these transformer blocks within the same fundus modal to enhance the self-attention capability of each layer. 
The computational procedure for FFA and CFP token embeddings can be summarized as follows:
\begin{flalign}\label{eq2}
\begin{split}
    \mathbf{X}_{l}^{v} &= {\rm MSA(LN}(\mathbf{X}_{l-1}^{v})) + \mathbf{X}_{l-1}^{v} \\
    \mathbf{X}_{l}^{v} &= {\rm FFN(LN}(\mathbf{X}_{l}^{v})) + \mathbf{X}_{l}^{v}  \\
    \mathbf{Y}_{l}^{v} &= {\rm MSA(LN}(\mathbf{Y}_{l-1}^{v})) + \mathbf{Y}_{l-1}^{v} \\
    \mathbf{Y}_{l}^{v} &= {\rm FFN(LN}(\mathbf{Y}_{l}^{v})) + \mathbf{Y}_{l}^{v}  \\
\end{split}
\end{flalign}


Here, $l$ stands for the layer index and $l=1, \dots, L_{1}$. 
Through the computation described above, our model can learn the contextual relationship and extract features within each fundus view and fundus modal. 

\subsection{Multi-view fundus image fusion}
MFSWFM is proposed to fuse multi-view features and efficiently model long-range dependencies in fundus images of different views, as shown in \ref{Methods}(c). 
Specifically, we concatenate all view token embeddings of FFA images and add a multi-view position embedding to the concatenated token embedding to retain multi-view fundus positional information. 
Then, we reshape the token embeddings of each view into original spatial positions to provide absolute position awareness within multi-view fundus images. 
The computational procedure can be summarized as follows:

\begin{flalign}\label{eq3}
    \mathbf{X}_{0} & = {\rm R}\left( {\rm Concat}\left( \mathbf{X}_{L_{1}}^{1}, \cdots , \mathbf{X}_{L_{1}}^{V} \right) + \mathbf{X}_{mvpos} \right)\\
    & = \left[ \mathbf{X}_{mvpos}^{1}, \cdots , \mathbf{X}_{mvpos}^{V} \right]
\end{flalign}

where $\rm R(\cdot)$ is the reshape operation, $\rm Concat(\cdot)$ is the concatenate operation, $\mathbf{X}_{0}\in\mathbb{R}^{\sqrt{N} \times (V \cdot \sqrt{N})\times D}$ is the concatenated multi-view token embedding, $\mathbf{X}_{mvpos}^{v} \in \mathbb{R}^{\sqrt{N} \times \sqrt{N} \times D}$ is the token embeddings in $v$-th view after reshape operation, and $\mathbf{X}_{mvpos} \in \mathbb{R}^{(V \cdot N) \times D}$ is the multi-view position embeddings. 

MFSWFM conducts two successive transformer layers \cite{Swintransformer}, where the second layer can be regarded as a shifted version of the first layer. 
The computational procedure can be summarized as follows: 

\begin{flalign}\label{eq4}
\begin{split}
    \mathbf{X}_{l} &= {\rm W\text{-}MSA(LN}(\mathbf{X}_{l-1})) + \mathbf{X}_{l-1} \\
    \mathbf{X}_{l} &= {\rm FFN(LN}(\mathbf{X}_{l})) + \mathbf{X}_{l} \\
    \mathbf{X}_{l+1} &= {\rm SW\text{-}MSA(LN}(\mathbf{X}_{l})) + \mathbf{X}_{l} \\
    \mathbf{X}_{l+1} &= {\rm FFN(LN}(\mathbf{X}_{l+1})) + \mathbf{X}_{l+1} 
\end{split}
\end{flalign}
Here, $l$ stands for the layer index and $l$=$1$. 
As shown in \ref{figure_introduction2}, W-MSA in layer $l$ and SW-MSA in layer $l+1$ denote window-based multi-head self-attention using regular and shifted window partitioning configurations in multi-view fundus images, respectively. 
Finally, to preserve the positional relationship of multi-view fundus images, we average the token embedding at the same spatial position from each view by an average pooling operation for subsequent multi-modal fusion, as shown in \ref{Methods}(c). 

Compared to the multi-view fusion method \cite{huang2023model} and the multi-modal fusion method \cite{Transmed} using the regular self-attention mechanism, we employ the shifted window-based self-attention, which can reduce the computational complexity, in multi-view fundus images instead of global self-attention. 
Supposing each window contains $M \times M$ token embeddings, the computational complexity of an MSA and a W-MSA on the multi-view token embeddings is: 
\begin{flalign}\label{eq5}
    \Omega({\rm MSA}) = 4(VN)D^{2} + 2(VN)^{2}D, 
\end{flalign}
\begin{flalign}\label{eq6}
    \Omega({\rm W\text{-}MSA}) = 4(VN)D^{2} + 2M^{2}VND, 
\end{flalign}
where the former is quadratic to the token embedding number $VN$, and the latter is linear when $M$ is fixed. The computation of MSA is generally unaffordable for a large $VN$, while the MVM-MSA can model inter-view long-range dependencies with linear computing complexity. 
We omit SoftMax computation in determining complexity following \cite{Swintransformer}. 


\subsection{Multi-modal fundus image fusion}
The architecture of our proposed CFFT is presented in \ref{Methods2}(a), which maintains three multi-modal multi-scale cross-attention module (MMCAM) variants with different receptive field parameters $r$ and a concatenation operation. 
CFFT models the bidirectional correspondences between multi-scale features of target modal and query modal images and performs dual feature aggregation for both target and query. 
Following \cite{Shunted}, we reduce the receptive field of the model layer by layer, enabling it to mine richer multi-scale lesion information by setting $r$ = 8, $r$ = 4, and $r$ = 2, gradually. 
\ref{Methods2}(d) shows a schematic diagram of multi-scale fundus cross-attention receptive fields with different parameters $r$ in the proposed MMCAM. 
As shown in \ref{Methods2}(b), the MMCAM contains multiple LN layers, two Parallel MCA parts, and two FFN layers. 
As shown in \ref{Methods2}(c), our MCA is different from the standard cross-attention \cite{lin2021cat} in that we add a multi-scale window map (MSWM) block to aggregate multi-scale features and a local residual convolutional layer (LRCL) to complement local information. 
Specifically, given the modal 1 feature input $\hat{\mathbf{X}}$ and the modal 2 feature input $\hat{\mathbf{Y}}$. 
In the MCA computing process, one modal features tensor is projected into a query tensor, and another is projected into key and value tensors at first. 
The query $\mathbf{Q}$, key $\mathbf{K}$, and value $\mathbf{V}$ are mapped into different scale feature spaces for different heads indexed by $i$:
\begin{flalign}\label{eq8}
    \begin{split}
        \mathbf{Q}_{i} &= \hat{\mathbf{X}}\mathbf{W}_{i}^{\mathbf{Q}}, \\
        \mathbf{K}_{i}, \mathbf{V}_{i} &= {\rm MSWM}(\hat{\mathbf{Y}}, {\rm r}_{i})\mathbf{W}_{i}^{KV}, \\
        \mathbf{V}_{i} &= \mathbf{V}_{i} + {\rm LRCL}(\mathbf{V}_{i})
    \end{split}
    \end{flalign}
Here the $\rm MSWM(\cdot;r_{i})$ is the MSWM block in the $i$-th head with the rate of $r_{i}$. 
$\mathbf{W}_{i}^{Q}$ and $\mathbf{W}_{i}^{KV}$ are the parameters of the linear projection in the $i$-th head. 
Different from Shunted Transformer \cite{Shunted}, which uses large convolutional kernels to extract multi-scale features, we employ successive convolutional layers with small convolutional kernels. 
Each MSWM contains two branches with different receptive fields. 
When $r$ = 8, MSWM used three $2 \times 2$ convolutions with a stride of 2 to simulate a large-sized convolution with a receptive field of 8 as one branch, and two $2 \times 2$ convolutions with a stride of 2 to simulate a large-sized convolution with a receptive field of 4 as the other branch. 
When $r$ = 4, MSWM uses two $2 \times 2$ convolutions with a stride of 2 and one $2 \times 2$ convolution with a stride of 2. 
When $r$ = 2, MSWM uses a $2 \times 2$ convolution with a stride of 2 and a $1 \times 1$ convolution with a stride of 1. 
LRCL aims to extract local fundus features in each cross-attention layer and is composed of a $1 \times 3$ convolution, a $3 \times 1$ convolution, and a residual connection. 
Compared with the large-sized convolutional kernels, small kernel sizes can reduce computational complexity while providing an equal-sized receptive field. 

To allow cross-attention to consider both large-scale and small-scale features in the same layer, multiple attention heads are evenly divided through a split attention mechanism. Assuming the total number of heads is $h$, when $i \textless (h/2)$, MSWM outputs the result of the branch with a larger receptive field, and when $i \geq (h/2)$, MSWM outputs the result of the branch with a smaller receptive field. Through this setting, it is possible to change the focus on features of different scales by varying the index of different headers in the calculation of cross-attention weights at the same layer. 

Then, the MCA of $i$-th head is calculated by:

\begin{flalign}
    \begin{aligned}\label{eq9}
        \mathbf{h}_{i}  = \operatorname{softmax}\left(\frac{\mathbf{Q}_{i} \mathbf{K}_{i}^{T}}{\sqrt{D}}\right) \mathbf{V}_{i}
    \end{aligned} 
    \end{flalign}

In the last layer of CFFT, we employ a concatenation operation to obtain the final fusion multi-modal and multi-view feature vector. 

\subsection{Multi-task decoder}

\subsubsection{Retinopathy classification}
In retinopathy classification tasks, we employ a classifier as the decoder to identify the category of multi-modal and multi-view fundus images. 
We assume that there are $K$ classes in the retinopathy classification task. 
Given the final fusion feature from the encoder, we utilize an average pooling layer to compute the average visual vector $\mathbf{v}_{a} \in \mathbb{R}^{D}$. 
The classifier can be formulated as: 
\begin{flalign}\label{eq10}
    \mathbf{P}=\operatorname{softmax}(\mathbf{W}_{p} \cdot \mathbf{v}_{a}+b_{p})
\end{flalign}
where $\mathbf{p} = [p_{1}, \dots, p_{k}, \dots p_{K}]$ denotes the probability vector of pathology, in which $p_{k}$ is the probability of the subject belonging to the $i$-th class. $\mathbf{W}_{p}$ and $b_{p}$ are the weights and bias, respectively, of the fully connected layer. 
We select the class of largest value in $P$ as the prediction result. The encoder and decoder are jointly trained by minimizing the cross-entropy loss $\mathcal L_{CE}$:
\begin{flalign}\label{eq11}
    \mathcal L_{CE} = - \sum\limits_{k=1}\limits^{K}{\hat{p}_{k}ln{p_{k}}}
\end{flalign}
where $\hat{\mathbf{p}} = [\hat{p_{1}}, \dots, \hat{p_{k}}, \dots \hat{p_{K}}]$ is the ground truth. 

\subsubsection{Symptom report generation}
The symptom report generation tasks contain a multi-label retinopathy classification task and a findings generation task. 
In these tasks, the decoder contains the above classifier and a Re-LSTM. 
We design our Re-LSTM inspired by the architecture of hierarchical long short-term memory in related medical report generation works \cite{lan2021automatic, Co-Att}. 
Specifically, we compute the probability vector of impression prediction $\mathbf{p} = [p_{1}, \dots, p_{k}, \dots p_{K}]$ by using eq. \ref{eq10}. 
We select these classes $p_{k}$, in which $p_{k}>0.5$, as the impression. 
Similarly to eq. \ref{eq11}, we use cross-entropy loss $\mathcal L_{CE}$ as the classification loss function. 

The Re-LSTM contains a sentence LSTM and a word LSTM, in which the sentence LSTM generates topic vectors as the input of the word LSTM and the word LSTM predicts words for each sentence based on the topic vector. 
In the sentence LSTM at time step $s$, the average visual vector $\mathbf{v}_{a}$ and the hidden state $\mathbf{h}^{s}$ are used for generating the topic vector $\mathbf{t}^{s}$:
\begin{flalign}\label{eq12}
    \mathbf{h}_{sent}^{s} &= {\rm LSTM}(\mathbf{v}_{a} \oplus \mathbf{t}^{s-1}), \\
    \mathbf{t}^{s} &= {\rm tanh}((\mathbf{W}_{h}\mathbf{h}_{sent}^{s} + b_{h}) + (\mathbf{W}_{v}\mathbf{v}_{a} + b_{v}))
\end{flalign}
where ${\rm LSTM}$ is a 2-layer long short-term memory \cite{hochreiter1997long}. $\mathbf{W}_{h}$ and $\mathbf{W}_{v}$ are weight parameters. $b_{h}$ and $b_{v}$ are bias parameters. Then, a probability distribution over $\{end=1, continue=0\}$ is utilized to control the end of sentence generation. Similarly to \cite{lan2021automatic, Co-Att}, the topic vector $\mathbf{t}$ is used as the first input of the word LSTM. 
At each step, the word LSTM samples from a probability distribution over the words in the pre-processing vocabulary to generate words. The subsequent inputs are the words generated in the previous step. Finally, the final findings section is simply the concatenation of all the generated sequences after each word LSTM has generated its words. 

We employ a joint loss $\mathcal L$ to jointly train the encoder and decoder, where the training of the sentence LSTM and word LSTM is the combination of two cross-entropy loss: $\mathcal L_{sent}$ and $\mathcal L_{word}$. The entire model is trained by minimizing $\mathcal L$, which is defined as:
\begin{flalign}\label{eq13}
    \mathcal L &= \lambda_{1}\mathcal L_{CE} + \lambda_{2}\mathcal L_{S} + \lambda_{3}\mathcal L_{W}, \\
    \mathcal L_{S} &= \sum\limits_{s=1}\limits^{S}\mathcal L_{sent}(\mathbf{p}_{stop}^{s}, I\{s=S\}), \\
    \mathcal L_{W} &= \sum\limits_{s=1}\limits^{S}\sum\limits_{t=1}\limits^{T_{s}}\mathcal L_{word}(\mathbf{p}_{s}^{t}, \mathbf{y}_{s}^{t})
\end{flalign}
where $\lambda_{1}$, $\lambda_{2}$ and $\lambda_{3}$ are the loss weights. We set $\lambda_{1}=\lambda_{2}=\lambda_{3}=1$ in our experiments. $I$ is a sample, and $\mathbf{p}_{stop}^{s}$ denotes the probability distribution over $\{end=1, continue=0\}$. $\mathbf{p}_{s}^{t}$ and $\mathbf{y}_{s}^{t}$ denote the predicted word distribution and the ground truth of the $t$-th word in the $s$-th sentence. $T_{s}$ denotes the maximum length of $s$-th sentence, and $S$ denotes the maximum number of generated sentences. 

\section{Datasets and pre-processing}\label{dataset}
Due to the lack of publicly available CFP and FFA image pair datasets, we use two private multi-modal and multi-view fundus image datasets, including the single-label fundus pathology dataset (SFPD) and the multi-label symptom report dataset (MSRD), to evaluate the effectiveness of our methods. 
The effectiveness of SFPD and MSRD has been verified in previous works \cite{lan2021automatic,huang2023model}. 
These datasets do not contain any identifying marks and are not accompanied by text that might identify the individual concerned. 


\begin{table*}[htbp]
    \centering
    \caption{Quantitative distribution of 14 retinal-related disease categories in SFPD and MSRD, including Diabetic Retinopathy (DR), Macular Edema (ME), Age-related Macular Degeneration (AMD), Pathological Myopia (PM), Retinal Vein Occlusion (RVO), Central Serous Chorioretinopathy (CSC), Choroidal Neovascularisation (CNV), Macular Epiretinal Membrane (EM), Retinitis Pigmentosa (RP), Uveitis’ Disease (UD), Hypertensive Retinopathy (HR), Drusen, Normal and  Eales’ Disease (ED).}
      \begin{tabular}{cccccccccccccccc}
      \toprule
      \toprule
      \multicolumn{2}{c}{Class} & DR    & ME    & AMD   & PM    & RVO   & CSC   & CNV   & EM    & RP    & UD    & HR    & Drusen & Normal & ED \\
      \midrule
      \multirow{2}[3]{*}{Sample} & SFPD  & 1818  & 0     & 1304  & 670   & 532   & 395   & 0     & 201   & 96    & 0     & 88    & 0     & 70    & 50 \\
  \cmidrule{2-16}          & MSRD  & 1361  & 815   & 779   & 519   & 467   & 390   & 218   & 173   & 103   & 99    & 98    & 76    & 65    & 45 \\
      \bottomrule
      \bottomrule
      \end{tabular}%
    \label{datasets}%
  \end{table*}%

\begin{table*}[htbp]
    \centering
    \caption{The single-label retinopathy classification performance comparison between our method and reference methods. $Params$ and $Time$ are presented in millions ($M$) and seconds ($S$), respectively (mean ± std). }
    \resizebox{\linewidth}{!}{
        \begin{tabular}{l p{2.0cm}<{\centering} p{2.0cm}<{\centering} p{2.0cm}<{\centering} p{2.0cm}<{\centering} l l}
      \toprule
      \toprule
      \multicolumn{1}{l}{Model} & \multicolumn{1}{c}{$ACC (\%)$} & $F_{1} (\%)$ & $P (\%)$ & $R (\%)$ & \multicolumn{1}{c}{$Params (M)$} & $Time (S)$ \\
      \midrule
      \multicolumn{1}{l}{AlexNet \cite{AlexNet}} & \multicolumn{1}{c}{65.82±4.63} & 48.41±5.68 & 55.09±6.32 & 46.35±5.62 &   57.05 $\times$ 5    & 1.18 $\times$ 5 \\
      \multicolumn{1}{l}{VGG11 \cite{VGG}} & \multicolumn{1}{c}{74.52±4.23} & 56.86±5.94 & 60.36±7.02 & 56.02±6.45 &   128.81 $\times$ 5    & 1.57  $\times$ 5 \\
      \multicolumn{1}{l}{ResNet18 \cite{ResNet}} & \multicolumn{1}{c}{73.51±4.79} & 57.46±5.79 & 60.30±5.79 & 56.68±6.13 &    11.18 $\times$ 5   & 1.35  $\times$ 5 \\
      \multicolumn{1}{l}{ResNet34 \cite{ResNet}} & \multicolumn{1}{c}{74.24±4.58} & 59.09±6.98 & 60.48±6.22 & 57.76±6.23 &   21.29 $\times$ 5    & 1.40  $\times$ 5 \\
      \multicolumn{1}{l}{DenseNet121 \cite{DenseNet}} & \multicolumn{1}{c}{75.74±4.72} & 60.25±6.01 & 62.41±5.98 & 59.97±6.18 &   6.96 $\times$ 5    & 2.15  $\times$ 5 \\
      \multicolumn{1}{l}{ViT-6 \cite{Vit}} & \multicolumn{1}{c}{71.02±5.19} & 53.73±6.01 & 57.48±6.58 & 52.29±5.90 &    43.13 $\times$ 5   & 1.87  $\times$ 5 \\
      \multicolumn{1}{l}{ViT-12 \cite{Vit}} & \multicolumn{1}{c}{71.54±4.93} & 54.47±5.41 & 58.62±5.52 & 53.05±5.66 &   85.66 $\times$ 5  & 2.90  $\times$ 5 \\
      \multicolumn{1}{l}{NAT-B \cite{Nat}} & \multicolumn{1}{c}{77.43±4.64} & 61.62±6.13 & 65.79±5.62 & 60.05±6.53 &    89.67 $\times$ 5   & 3.34  $\times$ 5 \\
      \multicolumn{1}{l}{Shunted-B \cite{Shunted}} & \multicolumn{1}{c}{76.67±4.37} & 60.34±5.95 & 64.21±5.42 & 58.79±6.31 &    39.11 $\times$ 5   & 4.07  $\times$ 5 \\
      \multicolumn{1}{l}{Swin-B \cite{Swintransformer}} & \multicolumn{1}{c}{76.06±4.61} & 59.69±6.18 & 63.22±6.26 & 58.29±6.39 &   87.71 $\times$ 5    & 3.43  $\times$ 5 \\
      Vig \cite{Vig}   & \multicolumn{1}{c}{66.85±4.93} & 49.24±5.58 & 52.37±5.51 & 48.85±5.57 &    85.69 $\times$ 5   & 5.78  $\times$ 5 \\
      \midrule
      \multicolumn{1}{l}{P3D \cite{P3D}} & \multicolumn{1}{c}{63.97±4.71} & 43.75±3.50 & 48.87±5.12 & 43.24±3.29 &   24.95    & 3.97 \\
      \multicolumn{1}{l}{3D ResNet \cite{3DResNet}} & \multicolumn{1}{c}{63.08±8.20} & 45.20±5.73 & 47.04±6.66 & 45.17±5.49 &    33.18   & 3.40 \\
      \multicolumn{1}{l}{Su et al. \cite{Suetal}} & \multicolumn{1}{c}{68.07±4.80} & 50.12±5.24 & 53.87±6.30 & 48.66±4.96 &    55.91   & 2.80 \\
      \midrule
      \multicolumn{1}{l}{MVMFF-Net-152 \cite{lan2021automatic}} & \multicolumn{1}{c}{79.89±3.22} & 61.74±4.30 &   72.15±6.75    &    58.64±4.92   &   230.61    & 10.62 \\
      \multicolumn{1}{l}{MVMFF-Net-101 \cite{lan2021automatic}} & \multicolumn{1}{c}{78.46±2.20} & 58.86±3.81 &     69.19±7.73  &     55.84±4.29  &    199.33   & 8.31 \\
      \multicolumn{1}{l}{MVDRNet \cite{MVDRNet}} & \multicolumn{1}{c}{66.98±9.99} & 45.73±10.42 & 50.05±13.28 & 47.25±9.56 &    134.28   & 8.22 \\
      \multicolumn{1}{l}{MMEFM1 \cite{huang2023model}} & \multicolumn{1}{c}{77.16±3.72} & 62.58±3.26 & 65.00±4.45 & 62.11±2.95 &   171.32    & 13.53  \\
      \multicolumn{1}{l}{MMEFM2 \cite{huang2023model}} & \multicolumn{1}{c}{80.71±4.94} & 64.78±7.20 & 69.02±9.66 & 63.48±6.40 &    128.78   & 14.85  \\
      \multicolumn{1}{l}{Wang et al. \cite{wang2019two}} & \multicolumn{1}{c}{78.45±3.84} & 65.79±4.48 & 66.86±4.21 & 66.39±4.77 &    22.36 $\times$ 4   & 3.11 $\times$ 4 \\
      \multicolumn{1}{l}{TransMed \cite{Transmed}} & \multicolumn{1}{c}{70.10±7.09} & 50.04±7.69 & 56.16±5.20 & 52.33±9.83 &   110.25    & 12.14 \\
      \multicolumn{1}{l}{Mil-vt \cite{yu2021mil}} & \multicolumn{1}{c}{60.00±6.00} & 41.86±3.66 & 43.84±3.82 & 41.96±3.18 &   21.73 $\times$ 5    & 3.37 $\times$ 5 \\
      \multicolumn{1}{l}{Xiao et al. \cite{xiao2024cross}} & \multicolumn{1}{c}{78.57±3.14} & 64.17±3.20 & 67.72±4.37 & 62.71±2.84 &   177.80 $\times$ 4    & 8.99 $\times$ 4 \\
      \midrule
      Ours  & \multicolumn{1}{c}{\textbf{82.53±2.34}} & \textbf{69.37±3.79} & \textbf{72.44±4.30} & \textbf{68.02±3.50} &    142.35   & 7.91 \\
      \bottomrule
      \bottomrule
      \end{tabular}%
    }
    \label{single-label table}%
  \end{table*}%

\begin{table}[htbp]
    \centering
    \caption{The multi-label retinopathy classification performance comparison between our method and reference methods. }
    \scalebox{1.09}{
      \begin{tabular}{lccc}
      \toprule
      \toprule
      \multicolumn{1}{l}{Model} & \multicolumn{1}{r}{$MF_{1} (\%)$} & \multicolumn{1}{r}{$MP (\%)$} & \multicolumn{1}{r}{$MR (\%)$} \\
      \midrule
      ResNet-152-M \cite{lan2021automatic} & 64.01  & 65.28  & 62.80  \\
      TiNet \cite{TieNet} &    63.97   &   59.98    & 61.90 \\
      TiNet-M \cite{lan2021automatic} & 61.56  & 63.60  & 59.70  \\
      Co-Att \cite{Co-Att} &    59.51   &   53.94    & 56.57 \\
      Co-Att-M \cite{lan2021automatic} & 68.93  & 70.61  & 67.33  \\
      MMEFM1 \cite{huang2023model} & 66.39  & 68.69  & 64.24  \\
      MMEFM2 \cite{huang2023model} & 66.23  & 69.05  & 63.63  \\
      \midrule
      Ours  & \textbf{72.38}  & \textbf{75.00}  & \textbf{69.93}  \\
      \bottomrule
      \bottomrule
      \end{tabular}%
      }
    \label{multi-label table}%
  \end{table}%


\begin{table*}[htbp]
    \centering
    \caption{Comparison of the results obtained by our method and the compared methods for fundus image report generation. }
      \begin{tabular}{lccccccc}
      \toprule
      \toprule
      Model & BLEU-1 & BLEU-2 & BLEU-3 & BLEU-4 & METEOR & ROUGR & CIDEr \\
      \midrule
      CNN-RNN \cite{CNN-RNN} & 0.211  & 0.113  & 0.068  & 0.043  & 0.147  & 0.182  & 0.151  \\
      CNN-RNN-M \cite{lan2021automatic} & 0.279  & 0.171  & 0.119  & 0.089  & 0.214  & 0.239  & 0.185  \\
      LRCN \cite{LRCN} & 0.236  & 0.135  & 0.090  & 0.067  & 0.174  & 0.215  & 0.215  \\
      LRCN-M \cite{lan2021automatic} & 0.276  & 0.172  & 0.121  & 0.092  & 0.216  & 0.237  & 0.279  \\
      Soft-Att \cite{Soft-Att} & 0.308  & 0.198  & 0.146  & 0.114  & 0.237  & 0.280  & 0.305  \\
      Soft-Att-M \cite{lan2021automatic} & 0.315  & 0.216  & 0.167  & 0.140  & 0.253  & 0.293  & 0.328  \\
      TiNet \cite{TieNet} & 0.237  & 0.149  & 0.104  & 0.076  & 0.184  & 0.209  & 0.281  \\
      TiNet-M \cite{lan2021automatic} & 0.299  & 0.178  & 0.122  & 0.090  & 0.226  & 0.253  & 0.328  \\
      Co-Att \cite{Co-Att} & 0.301  & 0.226  & 0.172  & 0.135  & 0.317  & 0.326  & 0.383  \\
      Co-Att-M \cite{lan2021automatic} & 0.335  & 0.252  & 0.197  & 0.161  & 0.344  & 0.379  & 0.404  \\
      MVMFF-Net \cite{lan2021automatic} & 0.390  & 0.293  & 0.231  & 0.191  & 0.433  & 0.392  & 0.452  \\
      MMEFM 1 \cite{huang2023model} & 0.422  & 0.358  & 0.307  & 0.267  & 0.499  & 0.450  & 0.675  \\
      MMEFM 2 \cite{huang2023model} & 0.420  & 0.361  & 0.313  & 0.274  & 0.501  & 0.451  & 0.688  \\
      \midrule
      Ours  & \textbf{0.543} & \textbf{0.474} & \textbf{0.414} & \textbf{0.366} & \textbf{0.527} & \textbf{0.503} & \textbf{0.797} \\
      \bottomrule
      \end{tabular}%
    \label{report generation table}%
  \end{table*}%

SFPD is a proprietary single-label retinopathy classification dataset, consisting of 5,224 eye cases of patients with ten retinal-related disease categories. Details of ten targets are shown in \ref{datasets}. 
The sample partition of five-fold cross-validation is carried out in each category. 

MSRD is a proprietary fundus dataset for multi-label retinopathy classification and symptom report generation tasks, consisting of 3646 training samples, 520 verification samples, and 1042 testing samples. 
Each sample is composed of one Chinese symptom report and five different modalities of fundus photographs, as shown in \ref{figure_introduction1}. 
The pathologies in the samples can be classified into fourteen retinal-related pathology categories, and each sample has one or multiple categories. 
Details of fourteen targets in MSRD are shown in \ref{datasets}. 

In the SFPD and MSRD datasets, each sample is composed of five 512 × 512 pixel fundus photograph images which consist of one CFP image of the primary view and FFA images of four different views. 
For a fair comparison, we follow the preprocessing step as in \cite{huang2023model}, including image preprocessing, target annotating, and word tokenizing. 
In MSRD, we tokenize symptom reports to get a vocabulary, which has 755 words, filtering words with a frequency of no more than two. 
In our experiments, the value of the total view number $V$ is set to four according to the clinical experience of the ophthalmologist. 

\begin{figure*}[htbp]
	\centering
	\begin{minipage}{0.49\linewidth}
		\centering
		\includegraphics[width=1.0\linewidth]{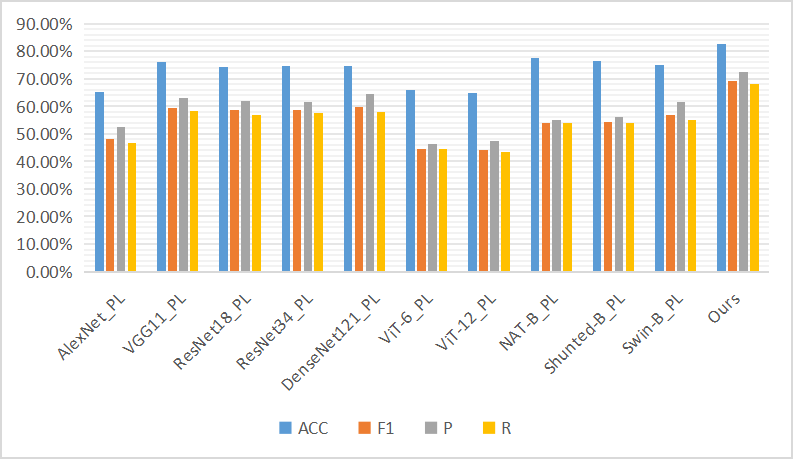}
		\centerline{(a)}
	\end{minipage}
	\begin{minipage}{0.49\linewidth}
		\centering
		\includegraphics[width=1.0\linewidth]{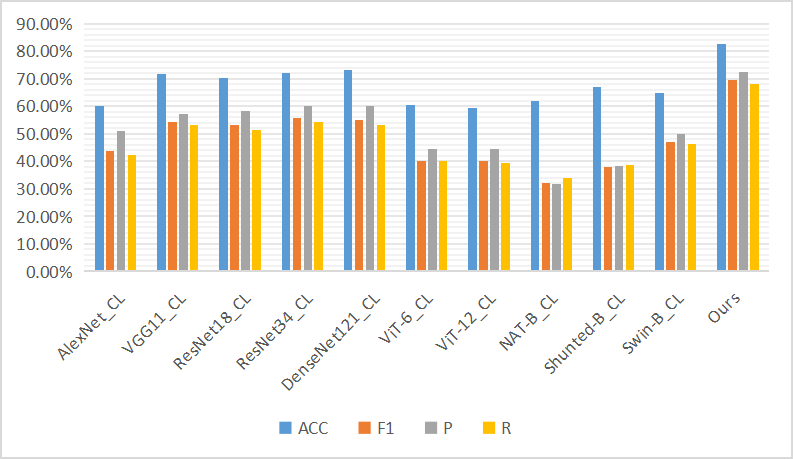}
		\centerline{(b)}
	\end{minipage}
    \caption{(a) The $ACC$, $F_{1}$, $P$, $R$ for single-label retinopathy classification of our model and the compared multi-modal fusion methods merging fundus images at the pixel level; 
    (b) The $ACC$, $F_{1}$, $P$, $R$ for single-label retinopathy classification of our model and the compared multi-modal fusion methods merging fundus images at the channel level. }
    \label{pixel_channel}
\end{figure*}


\begin{figure*}[htb]
    \begin{center}
        \includegraphics[width=1.00\textwidth]{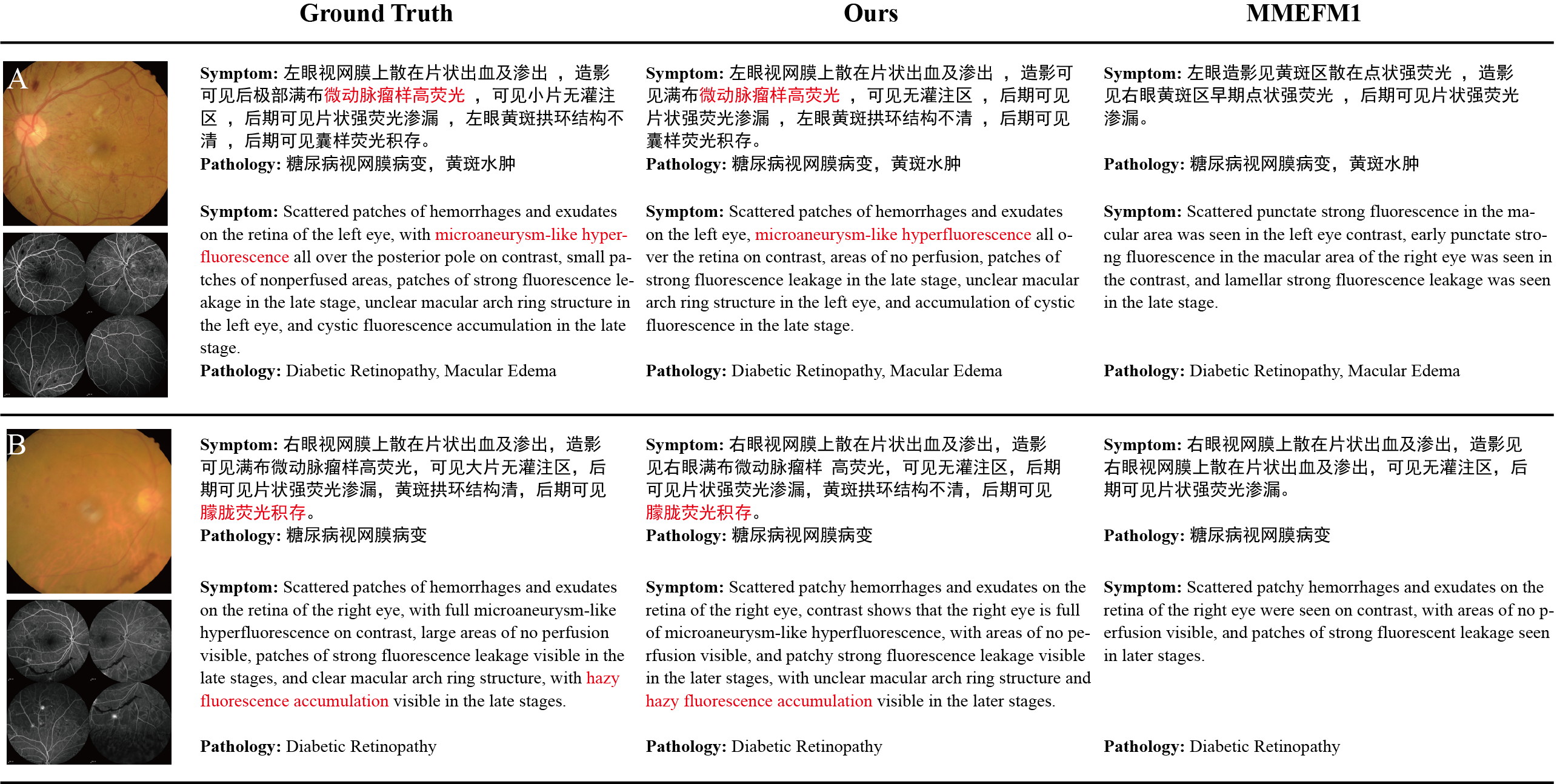}
        \caption{Examples of reference ground truth fundus reports and fundus reports obtained by our method and MMEFM1 \cite{huang2023model}. The key findings are marked in red. The generated results are translated from Chinese to English for display purposes. } 
        \label{results} 
    \end{center}
\end{figure*} 

\section{Experiments and results}\label{experiments and results}
\subsection{Experiment setup for retinopathy classification} 
\label{classification}

To balance the computational complexity and classification accuracy, each image is resized to 224 × 224 for both datasets. 
We set $M=7$ and $P=16$ in our experiments. 
We set $L_{1}=6$ and load the pre-train weights ViT-Base \cite{Vit} with 6 layers in the backbone. 
We set $h=16$ in this work. 

All experiments are conducted on an NVIDIA RTX 3090 GPU. All models are trained for 100 epochs with the Adam optimizer \cite{kingma2014adam}. 
A batch size of 16 and an initial learning rate of 0.0001 are used. 
We use a learning rate dynamic adjustment strategy based on the loss function. 
On these two datasets, we do data augmentation strategies ambitiously, including random horizontal flip, random vertical flip and random rotation during training. 
The probability of each strategy being used is 0.5. 
To compare the performance of the single-label classification on SFPD, we compute four metrics, the $Accuracy$ $(ACC)$, $Precision$ $(P)$, $Recall$ $(R)$ and $F_{1}$-$Score$ $(F_{1})$. 
For the multi-label classification tasks on MSRD, multi-label metrics \cite{zhang2013review}, including the $macro$-$averaged$ $F_{1}$-$Score$ $(MF_{1})$, the $macro$-$averaged$ $Recall$ $(MR)$ and the $macro$-$averaged$ $Precision$ $(MP)$ are leveraged to evaluate model performance for multi-label retinopathy classification. 
Additionally, model parameters ($Params$) and inference time ($Time$) are utilized to demonstrate the complexity of various methods. 

\begin{table*}[htbp]
    \centering
    \caption{Ablation experiment of MFSWFM and CFFT on the SFPD and the MSRD datasets performance (mean ± std). }
      \begin{tabular}{lcc|cccc|ccc}
      \toprule
      \multicolumn{1}{r}{\multirow{2}[4]{*}{Model}} & \multirow{2}[4]{*}{MFSWFM} & \multirow{2}[4]{*}{CFFT} & \multicolumn{4}{c|}{SFPD}     & \multicolumn{3}{c}{MSRD} \\
  \cmidrule{4-10}          &       &       & $ACC (\%)$   & $F_{1} (\%)$    & $P (\%)$     & $R (\%)$     & $MF_{1} (\%)$   & $MP (\%)$    & $MR (\%)$ \\
      \midrule
            & \textbf{\checkmark} &       & 79.74±4.12 & 64.96±4.37 & 67.82±4.28 & 63.83±4.28 & 66.74 & 69.87 & 63.87 \\
            &       & \textbf{\checkmark} & 79.22±2.95 & 64.24±3.85 & 66.63±4.93 & 63.56±3.39 & 66.46 & 68.83 & 64.25 \\
      Ours  & \textbf{\checkmark} & \textbf{\checkmark} & \textbf{82.53±2.34} & \textbf{69.37±3.79} & \textbf{72.44±4.30} & \textbf{68.02±3.50} & \textbf{72.38} & \textbf{75.00} & \textbf{69.93} \\
      \bottomrule
      \end{tabular}%
    \label{MFSWFMandCFFT}%
  \end{table*}%

\begin{table*}[htbp]
    \centering
    \caption{Ablation experiment of LRCL on the SFPD dataset performance (mean ± std). w/o means without, w means with. }
      \begin{tabular}{lcccc}
      \toprule
      Model & $ACC (\%)$ & $F_{1} (\%)$ & $P (\%)$ & $R (\%)$ \\
      \midrule
      MRDF w/o LRCL & 78.77±5.07 & 64.62±4.26 & 66.92±5.47 & 64.26±2.74 \\
      MRDF w LRCL & \textbf{82.53±2.34} & \textbf{69.37±3.79} & \textbf{72.44±4.30} & \textbf{68.02±3.50} \\
      \bottomrule
      \end{tabular}%
    \label{LRCL}%
  \end{table*}%
  
\begin{table*}[htbp]
    \centering
    \caption{Ablation experiment of position embeddings and multi-view position embeddings on the SFPD dataset performance (mean ± std). }
      \begin{tabular}{lcccccc}
      \toprule
      Model & position embeddings   & multi-view position embeddings & $ACC (\%)$   & $F_{1} (\%)$    & $P (\%)$     & $R (\%)$ \\
      \midrule
            & \textbf{\checkmark}     &       & 75.23±5.21 & 59.80±4.37 & 62.77±4.93 & 59.43±3.29 \\
            &       & \textbf{\checkmark}     & 80.89±1.76 & 66.94±2.38 & 69.61±4.06 & 66.43±3.05 \\
      Ours  & \textbf{\checkmark}     & \textbf{\checkmark}     & \textbf{82.53±2.34} & \textbf{69.37±3.79} & \textbf{72.44±4.30} & \textbf{68.02±3.50} \\
      \bottomrule
      \end{tabular}%
    \label{pos}%
  \end{table*}%

\begin{table*}[htbp]
    \centering
    \caption{Ablation experiment of $r$ in CFFT on the SFPD dataset performance (mean ± std). }
      \begin{tabular}{cccccccc}
      \toprule
      Model & $r=8$   & $r=4$   & $r=2$   & $ACC (\%)$   & $F_{1} (\%)$    & $P (\%)$     & $R (\%)$ \\
      \midrule
            & \textbf{\checkmark}     & \textbf{\checkmark}     &       & 81.85±2.01 & 68.01±4.20 & 71.07±3.12 & 67.66±5.67 \\
            & \textbf{\checkmark}     &       & \textbf{\checkmark}     & 80.94±3.56 & 67.30±3.58 & 71.55±3.98 & 65.79±3.22 \\
            &       & \textbf{\checkmark}     & \textbf{\checkmark}     & 81.22±3.43 & 68.22±3.78 & 71.04±4.60 & 67.44±3.51 \\
            & \textbf{\checkmark}     &       &       & 80.99±2.45 & 63.78±4.57 & 66.29±4.39 & 63.11±4.95 \\
            &       & \textbf{\checkmark}     &       & 81.23±2.37 & 65.21±3.88 & 67.79±5.43 & 64.59±3.78 \\
            &       &       & \textbf{\checkmark}     & 81.22±1.96 & 66.55±6.22 & 68.44±6.09 & 66.16±7.20 \\
      Ours  & \textbf{\checkmark}     & \textbf{\checkmark}     & \textbf{\checkmark}     & \textbf{82.53±2.34} & \textbf{69.37±3.79} & \textbf{72.44±4.30} & \textbf{68.02±3.50} \\
      \bottomrule
      \end{tabular}%
    \label{r}%
  \end{table*}%

We compare our method with multi-modal and multi-view medical classification methods, including the model from \cite{xiao2024cross}, the model from \cite{huang2023model}, MVMFF-Net \cite{lan2021automatic}, MVDRNet \cite{MVDRNet}, the model from \cite{wang2019two} and a method based on magnetic resonance imaging \cite{Transmed}. 
We also compare our model with the multi-modal fusion method based on the input-level fusion strategy like P3D \cite{P3D}, 3D ResNet \cite{3DResNet} and the model form \cite{Suetal}. 
In addition, some existing state-of-the-art single-modal methods are also included in the comparison, including CNN-based methods \cite{AlexNet,VGG,ResNet,DenseNet} and transformers-based methods \cite{yu2021mil,Vit,Nat,Shunted,Swintransformer,Vig}. 
For the sake of objectivity in the experiment, inspired by \cite{li2024review}, we implemented the decision-level fusion strategy in the above single-modal methods and a single-modal fundus image classification method \cite{yu2021mil}. 
Besides, we also compare our model with the multi-modal fusion methods based on the input-level fusion strategy \cite{li2024review} that merges data by adding at the pixel level and concatenating at the channel level. 
Specifically, we use the above single-modal methods as the basic structure, and we call these methods merging data at pixel level ViT-6\_PL, etc and these methods merging data at channel level ViT-6\_CL, etc. 
We combine a CFP and an FFA image as input to the models from \cite{wang2019two} and \cite{xiao2024cross}, since these models accept input from two modalities. 
In the MSRD, the multi-label classification task and the report generation task are related. 
Following previous related works \cite{huang2023model,lan2021automatic}, we choose some existing medical report generation methods including the model from \cite{huang2023model}, these methods from \cite{lan2021automatic}, TieNet \cite{TieNet}, Co-Att \cite{Co-Att} as comparisons. 

\subsection{Experimental results on retinopathy classification}
\ref{single-label table} presents the single-label retinopathy classification results of our mothod and multi-modal and multi-view fusion methods. 
We have the following observations: 1) Our proposed method outperforms the decision-level multi-modal fusion methods \cite{li2024review} using the single-modal models as the backbone. 
This demonstrates that our multi-modal and multi-view fusion strategy can aggregate more vital information of different modalities and views than the previous decision-level multi-modal fusion strategy. 
2) Specially designed multi-modal and multi-view fusion methods, such as the model from Xiao et al. \cite{xiao2024cross} and MVMFF-Net \cite{lan2021automatic}, achieve higher classification metrics compared to the decision-level multi-modal fusion methods \cite{li2024review} using single-modal models in most scenarios. 
One potential reason is that the decision-level multi-modal fusion methods cannot fully use complementary information between different modalities and perspectives. 
Additionally, our method achieves higher classification results in all metrics than these specially designed multi-modal and multi-view fusion methods. 
These results indicate our model's strong information fusion capacity for accurately diagnosing retinopathy in multi-modal and multi-view CFP and FFA images. 
3) Although 3D methods based on input-level fusion strategies have fewer model parameters and shorter inference times than other methods in most scenarios, their classification results are worse in most cases. 
One potential reason is that these methods of processing multi-modal and multi-view fundus images by concatenating them together into 3D data ignore the differences in semantic information between different modalities and perspectives, which leads to a decrease in model performance. 
\ref{pixel_channel} (a) and \ref{pixel_channel} (b) show the $ACC$, $P$, $R$, $F_{1}$ for single-label retinopathy classification results of our method and the compared multi-modal fusion methods merging fundus images at the pixel level, 
and the results of our method and the compared multi-modal fusion methods merging fundus images at the channel level, respectively. 
The results in \ref{pixel_channel} demonstrate that our model substantially outperformed all multi-modal input-level fusion methods based on the pixel level and channel level regarding $ACC$, $P$, $R$, and $F_{1}$ metrics. 
It further verifies the effectiveness of our proposed multi-modal and multi-view fundus fusion strategy based on multi-scale cross-attention and shifted window self-attention. 
\ref{multi-label table} shows the multi-label retinopathy classification results obtained by our method alongside the compared methods. 
Our method surpassed the second-best performer by 3.45\% in $MF_{1}$, 4.39\% in $MP$, 2.60\% in $MR$. 
Overall, these results demonstrate the effectiveness of the fusion method we proposed and its generalization ability in different retinal disease diagnosis tasks.
We compare the computational efficiency of our models with other models in terms of the number of parameters and inference time, as shown in \ref{single-label table}. 
Most comparison methods contain more network parameters and more inference time than our proposed models. 
For instance, our model reduces inference time by almost half compared to these multi-modal and multi-view fusion methods based on regular self-attention \cite{huang2023model}. 
These results demonstrate the effectiveness of our multi-modal and multi-view fusion method based on the shifted window self-attention. 
Compared with methods with fewer parameters than our proposed models, our proposed models achieve significant performance improvement and almost the same inference speed. 

\subsection{Experiment setup for symptom report generation}

All fundus image are resized to 224 × 224. 
We use the same data enhancement strategies as in the classification experiments.
All models are trained for 50 epochs. 
A batch size of 8 and an initial learning rate of 0.0001 are used. 
We set the size of hidden units and word embeddings in the multi-task decoder to 256. 
We initialize the hidden unit and sentence composition vectors using a randomization function. 
Set the maximum number of words and sentences in the generated report to 20 and 10, respectively. 
The experimental environment and the parameters of our model are consistent with that in classification experiments. 
Four common report generation metrics are selected for evaluation of generated report results, including BLEU \cite{BLEU}, ROUGE \cite{ROUGE}, METEOR \cite{METEOR}, and CIDEr \cite{CIDEr}. 
We set BLEU-n paradigms of grams to four. 

\subsection{Experimental results on symptom report generation}
To the best of our knowledge, there are only a few previous
literature for fundus report generation using multi-modal multi-view fundus images. 
For the fundus report generation task, we employ some existing medical report generation methods, including models form \cite{lan2021automatic}, the model from \cite{huang2023model}, CNN-RNN \cite{CNN-RNN}, LRCN \cite{LRCN}, Soft-Att \cite{Soft-Att}, TieNet \cite{TieNet}, Co-Att \cite{Co-Att}. 
For the sake of objectivity in the comparison experiment, we follow prior works \cite{huang2023model,lan2021automatic} to show the results of these methods. 
The fundus report generation results are reported in \ref{report generation table}. 
Our method's report generation results exceeded the second-highest metrics by 0.121 in BLEU-1, 0.113 in BLEU-2, 0.101 in BLEU-3, 0.092 in BLEU-4, 0.026 in METEOR, 0.052 in ROUGR, and 0.109 in CIDEr. 
As shown in \ref{results}, we show two generated fundus reports with our method and MMEFM1 \cite{huang2023model}. 
For case A, compared to MMEFM1 disregarded some important ophthalmic symptoms, such as microaneurysm-like hyperfluorescence, our method can generate complete ophthalmic symptom paragraphs. 
Additionally, the generated report from our method can contain coherent ophthalmic symptom contexts such as hazy fluorescence accumulation for case B, while MMEFM1 failed to generate them. 

\section{Ablation study}
\subsection{Selection of essential components}
MRDF has two vital fusion components: MFSWFM for multi-view fundus image fusion and CFFT for multi-modal fundus image fusion. 
Therefore, we perform some ablation experiments on the SFPD and MSRD datasets with and without these components while keeping other settings constant, as shown in \ref{MFSWFMandCFFT}. 
The performance of our methods achieves improvements in all metrics compared to the performance without MFSWFM or CFFT. 
These results indicate the effectiveness of using MFSWFM and CFFT to fuse features of different modalities and views. 

We then conducted an ablation study on the LRCL of MCA to analyze its contributions. As shown in \ref{LRCL}, removing the LRCL resulted in decreases across all classification metrics on SFPD. Specifically, the $ACC$, $F_{1}$, $P$, and $R$ were decreased by 3.76\%, 4.75\%, 5.52\%, and 3.76\%, respectively. 
The results indicate the effectiveness of using LRCL to extract local features in each cross-attention layer. 

\subsection{Selection of position embeddings}
We then conducted an ablation study on the position embeddings and multi-view position embeddings to analyze their contributions to retaining spatial location information in multi-modal and multi-view fundus images, as shown in \ref{pos}. 
When we removed the position embeddings, $ACC$, $F_{1}$, $P$, and $R$ were decreased by 1.64\%, 2.43\%, 2.83\%, and 1.59\%, respectively. 
When we removed the multi-view position embeddings, $ACC$, $F_{1}$, $P$, and $R$ were decreased by 7.30\%, 9.57\%, 9.67\%, and 8.59\%, respectively. 
These results show that the position embeddings and the multi-view position embeddings play essential roles in providing absolute position awareness and capturing relative positional relationships between different views in improving our model's retinopathy diagnosis performance. 

\subsection{Analysis of parameters in CFFT}
We also conducted experiments to evaluate the effects of different $r$ in CFFT on the SFPD dataset, as shown in \ref{r}. 
When we removed any $r$ in CFFT, all the metric values declined. 
For instance, when we removed the $r=8$ and $r=4$, $ACC$, $F_{1}$, $P$, and $R$ were decreased by 1.31\%, 2.82\%, 4.00\%, and 1.86\%, respectively. 
When we removed the $r=2$, $ACC$, $F_{1}$, $P$, and $R$ were decreased by 0.68\%, 1.36\%, 1.37\%, and 0.36\%, respectively. 
These results show that each value of the parameter $r$ plays a vital role in providing different scale receptive fields and improving our model's performance. 

\section{Conclusion}\label{conclusions}
In this paper, we propose a multi-modal and multi-view fundus image multi-tasks framework that can simultaneously process fundus images of different views and modalities for retinopathy pathological classification and symptom report generation tasks. 
The proposed framework can handle any number of modalities and views by expanding the transformer block in the backbone. 
In this work, a multi-view fundus image fusion method called MFSWF can efficiently capture comprehensive fundus fields and lesion areas in multi-view fundus images. 
To mine the correspondence relationship between coarse-grained and fine-grained lesion features of multi-modal fundus images, we propose a multi-modal fundus fusion method called CFFT, which has a dual-branch structure with multiple different scale receptive fields. 
Therefore, our work holds great potential to support ophthalmologists in analyzing multi-modal multi-view fundus images and completing fundus symptom reports in clinical practice. 
In future works, we also plan to reduce our model's complexity and extend our model to 3D multi-modal medical imaging datasets. 



\section*{Acknowledgements}
This study was supported by Natural Science Foundation of  Sichuan, China(No. 2023NSFSC0468, No. 2023NSFSC0031)












\printcredits

\bibliography{cas-dc-template.bib}




\end{document}